\documentclass[letterpaper, 10 pt, conference]{ieeeconf}  

\IEEEoverridecommandlockouts                              

\usepackage{cite}
\usepackage{amsmath,amssymb,amsfonts}
\usepackage[e]{esvect}
\usepackage{graphicx}
\usepackage{siunitx} 

\usepackage{caption}

\usepackage{subcaption}
\usepackage{textcomp}
\usepackage{color}
\usepackage[colorlinks,
 linkcolor=black,
 anchorcolor=blue,
 citecolor=blue
 ]{hyperref} 
\usepackage[noabbrev]{cleveref}
\usepackage{multirow}
\usepackage{tabularray}
\usepackage[T1]{fontenc}
\usepackage{algorithm} 
\usepackage[noend]{algpseudocode}
\usepackage{fancyhdr}
\usepackage{dsfont}
\usepackage{censor}

\usepackage{url}
\usepackage{multicol}
\usepackage{multirow}
\usepackage{booktabs}

\usepackage{stfloats}
\usepackage{makecell}
\usepackage{easyReview}

\usepackage{xspace}

\usepackage[acronym]{glossaries}
\newacronym{rl}{RL}{Reinforcement Learning}
\newacronym{mri}{MRI}{magnetic resonance imaging}
\newacronym{ct}{CT}{Computed Tomography}
\newacronym{us}{US}{ultrasound}
\newacronym{nir}{NIR}{near-infrared}
\newacronym{icg}{ICG}{indocyanine green}
\newacronym{sbr}{SBR}{signal-to-background ratio}
\newacronym{snr}{SNR}{signal-to-noise ratio}
\newacronym{rapn}{RAPN}{robot-assisted partial nephrectomy}
\newacronym{dvrk}{dVRK}{da Vinci Research Kit}
\newacronym{pam}{PAM}{polyacrylamide}
\newacronym{pva}{PVA}{polyvinyl alcohol}
\newacronym{pla}{PLA}{polylactic acid}
\newacronym{dsc}{DSC}{DICE similarity coefficient}
\newacronym{mis}{MIS}{minimally-invasive surgery}
\newacronym{our_pipeline}{OA-NBV}{Occlusion-Aware Next-Best-View Planning for Human-Centered Active Perception on Mobile Robots}
\newacronym{nbv}{NBV}{Next-Best-View}
\newacronym{hpe}{HPE}{Human Pose Estimation}
\newacronym{nerf}{NeRF}{Neural Radiance Field}
\newacronym{prednbv}{Pred-NBV}{Pred-NBV}
\newacronym{volnbv}{Volumetric-NBV}{Volumetric-NBV}

\usepackage[table]{xcolor}
\definecolor{pastelblue}{RGB}{173,216,230}
\definecolor{pastelpink}{RGB}{255,182,193}
\definecolor{pastelyellow}{RGB}{255,245,170}
\definecolor{pastelpurple}{RGB}{216,191,216}
\definecolor{headergray}{RGB}{235,235,235}
\definecolor{initial_camera_view}{RGB}{249,203,223}
\definecolor{best_camera_view}{RGB}{204,231,207}
\definecolor{spare}{RGB}{245,235,217}
\usepackage{array}
\usepackage{multirow}
\newcommand{\cellw}{0.245\linewidth}
\newcommand{\cellh}{0.18\linewidth}
\newcommand{\cellbox}[1]{%
  \raisebox{0pt}[\cellh][0pt]{%
    \includegraphics[width=\cellw,height=\cellh,keepaspectratio]{#1}%
  }%
}
\usepackage{graphicx}
\usepackage{array}
\usepackage[table]{xcolor}
\usepackage{adjustbox}

\usepackage{titlesec}
\titlespacing*{\section}{0pt}{*0.5}{*0.5}
\titlespacing*{\subsection}{0pt}{*0.4}{*0.4}

\title{\LARGE \bf OA-NBV: Occlusion-Aware Next-Best-View Planning for Human-Centered Active Perception on Mobile Robots
 \vspace{-20pt}}

\author{
Boxun Hu$^{a}$,
Chang Chang$^{a}$,
Jiawei Ge$^{b}$,
Man Namgung$^{c}$,
Xiaomin Lin$^{a,d}$,
Axel Krieger$^{b}$,
and Tinoosh Mohsenin$^{a}$
\\
$^{a}$Department of Electrical and Computer Engineering,
Johns Hopkins University, Baltimore, MD 21218, USA\\
$^{b}$Department of Mechanical Engineering,
Johns Hopkins University, Baltimore, MD 21218, USA\\
$^{c}$Laboratory for Computational Sensing and Robotics,
Johns Hopkins University, Baltimore, MD 21218, USA\\
$^{d}$Department of Electrical Engineering,
University of South Florida, Tampa, FL 33620, USA\\
E-mail: \{bhu29, cchan144, jge9, mnamgun1, axel, Tinoosh\}@jhu.edu;
xlin2@usf.edu 
}
\begin{document}
\twocolumn[{
  \renewcommand\twocolumn[1][]{#1} 
  \maketitle
  \vspace{-18pt}
  \centering
  \includegraphics[width=\textwidth]{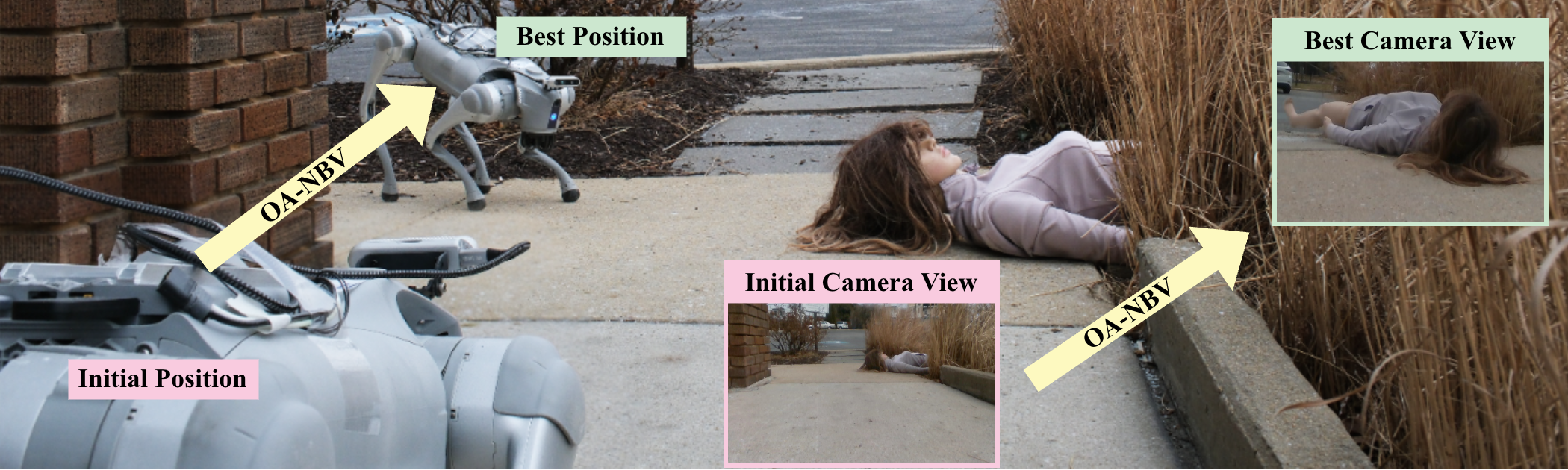}
  \captionof{figure}{\textbf{
    Occlusion-Aware Next-Best-View (OA-NBV) Planning for Human-Centered Active Perception on Mobile Robots.} 
    Given an RGB image of a mannequin in an outdoor scene and a point cloud observed from the current viewpoint (pink), \glsentryshort{our_pipeline} identifies the next-best viewpoint (green) that maximizes target visibility under terrain traversability constraints.
}
  \label{fig:01}
}]
\begin{abstract}

    We naturally step sideways or lean to see around the obstacle when our view is blocked, recovering a more informative observation. In disaster response, where medics and robot operators are often limited, enabling robots to make the same kind of viewpoint choice without continuous human guidance is critical for human-centered operations such as search and triage, especially when cluttered environments and partial visibility degrade downstream perception. However, many \gls{nbv} methods primarily optimize generic exploration or long-horizon coverage, and do not explicitly target the immediate goal of obtaining a single usable observation of a partially occluded person under real motion constraints.
    We present \gls{our_pipeline}, an occlusion-aware NBV pipeline that autonomously selects the next traversable viewpoint to obtain a more complete view of an occluded human. \gls{our_pipeline} integrates perception and motion planning by scoring candidate viewpoints using a target-centric visibility model that accounts for occlusion, target scale, and target completeness, while restricting candidates to feasible robot poses.
    \gls{our_pipeline} achieves over \SI{90}{\percent} success rate in both simulation and real-world trials, while baseline NBV methods degrade sharply under occlusion.
    Beyond success rate, \gls{our_pipeline} improves observation quality: compared to the strongest baseline, it increases normalized target area by at least \SI{81}{\percent} and keypoint visibility by at least \SI{58}{\percent} across settings, making it a drop-in view-selection module for diverse human-centered downstream tasks.  \vspace{-10pt}

\end{abstract}
\section{Introduction}
In cluttered, unstructured environments, a mobile robot rarely observes a person from a clean, front-facing view: debris, furniture, and scene geometry can occlude critical body regions, making perception strongly viewpoint-dependent. Humans handle this effortlessly by taking a small step or leaning to see around an occluder, quickly recovering an informative view~\cite{ballard1991animate}. For robots operating in human-centered settings such as search, triage, and disaster response~\cite{murphy2017disaster}, an analogous capability is essential because occlusion makes perception sensitive to viewpoint changes. In these settings, downstream reasoning benefits from seeing more of the body. A more complete view provides posture and limb cues and reduces ambiguity, supporting more reliable triage decisions in scenarios where medical personnel are limited. When only small regions are visible, partial observations can break person detection~\cite{zhang2017citypersons}, keypoint estimation~\cite{jiang2023rtmpose}, and 3D human reconstruction~\cite{kolotouros2019learning}. These failures can propagate to higher-level decisions~\cite{kiciroglu2020activemocap}.

This problem falls under active perception~\cite{bajcsy2018revisiting}: the robot must select the \gls{nbv} that produces an immediately usable observation of the occluded target, rather than optimizing for broad exploration. However, many NBV methods were designed around objectives such as scene/volume coverage~\cite{bircher2016receding}, uncertainty reduction~\cite{vasquez2017view}, or long-horizon reconstruction quality~\cite{chen2024gennbv, shen2025auto3r}, which do not necessarily align with short-horizon perceptual reliability for a partially occluded person. In clutter, a viewpoint that scores well under a generic exploration objective may still be blocked by foreground occluders~\cite{xue2024neural}, and viewpoint sets that ignore kinodynamic and collision constraints can be infeasible or unsafe on real mobile platforms~\cite{bircher2016receding, tabib2016computationally}.

We address this problem with \gls{our_pipeline} as shown in Fig.~\ref{fig:01}, an occlusion-aware NBV pipeline for mobile robots that explicitly optimizes for obtaining a single, high-quality observation of a partially occluded target under real motion constraints. \gls{our_pipeline} closes the loop between perception and motion in two stages. First, it extracts target 3D information from the current RGB image and point cloud to form a geometric hypothesis of the person and the visible region. Second, it generates traversable pose candidates and scores them using a target-centric visibility model that accounts for environmental occlusion, target scale, and target completeness, selecting the next viewpoint that is both informative and reachable. 
We validate \gls{our_pipeline} in realistic simulation environments with severe occlusions and through real-world deployment on a quadruped robot in indoor and outdoor scenes. Across iterative NBV steps, \gls{our_pipeline} maintains high success rates under occlusion and produces more complete target observations than representative volumetric and prediction-guided baselines~\cite{aleotti2014global,dhami2023pred}, improving both observation reliability and quality.
Our contributions are threefold:\vspace{-5pt}
\begin{itemize}
    \item \textbf{Occlusion-aware viewpoint scoring:} A target-centric visibility model that jointly accounts for occlusion, target scale, and in-frame completeness to select viewpoints for human-centered perception.
    \item \textbf{Part-aware 3D target estimation:} A pipeline that reconstructs human geometry from partial observations using segmentation-guided 3D lifting and part-aware mesh-to-point-cloud alignment, enabling robust target localization under severe occlusion.
    \item \textbf{Traversability-constrained viewpoint generation:} An elevation-map-based viewpoint sampling strategy that respects terrain traversability and robot kinematics, ensuring all candidate viewpoints are physically reachable in cluttered environments.
\end{itemize}
\section{Related Work}

\begin{figure*}[!t]
\centering
\includegraphics[width=\linewidth]{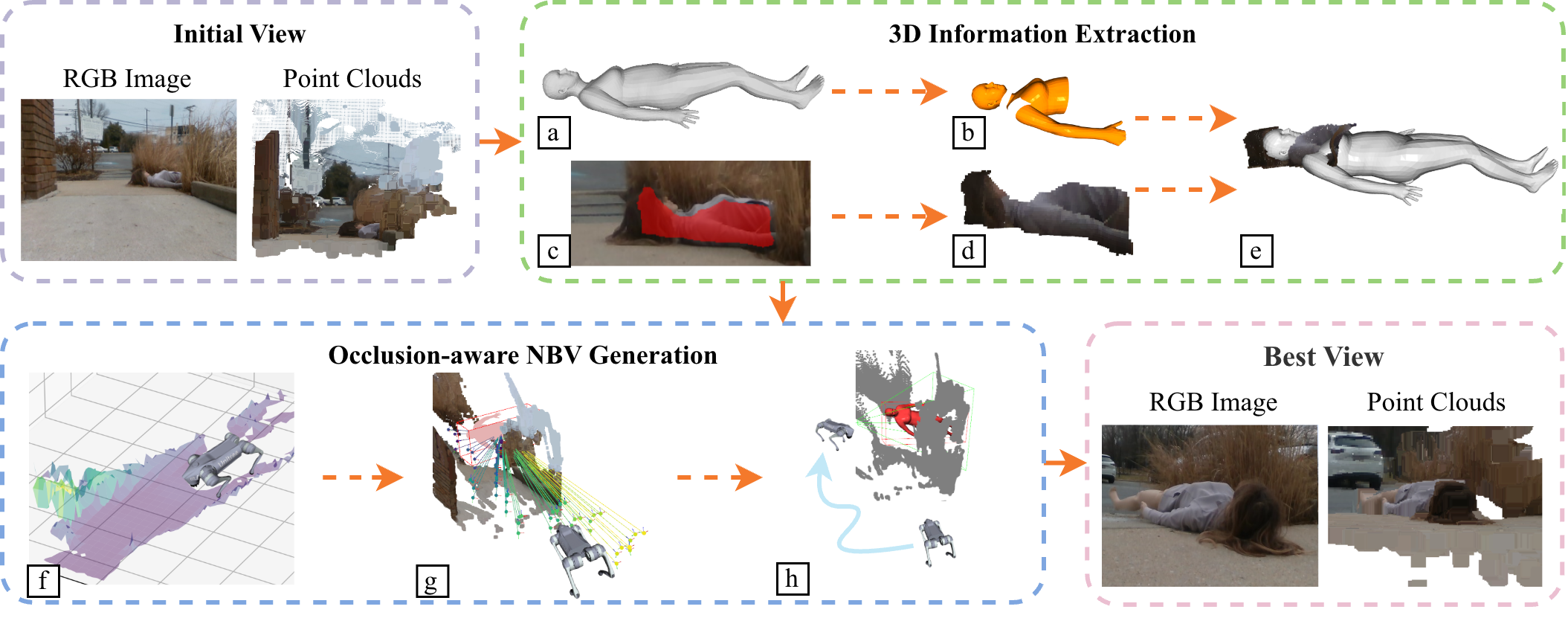}
\caption{\textbf{Overview of \gls{our_pipeline}.}
Given an RGB image and its paired point cloud from the initial viewpoint, 
\gls{our_pipeline} proceeds in two stages.
\textbf{3D Information Extraction:} 
(a) a human mesh is reconstructed from the RGB input,
(b) body part meshes are obtained according to predicted labels, 
(c) a 2D segmentation mask is generated using SAM~2, and (d) projected onto the point cloud to isolate the 3D target region, 
(e) which is then registered with the detected body parts and mapped transformation back onto the complete mesh.
\textbf{Occlusion-aware NBV Generation:} 
(f) an elevation map is built from LiDAR measurements to model terrain geometry, 
(g) traversable pose candidates are sampled on the elevation map, 
and (h) the next-best viewpoint is selected from the candidates to maximize target visibility while respecting terrain constraints. 
The robot then navigates to the best viewpoint, acquiring an improved observation of the target. \vspace{-3pt}}
\label{fig:avo_overview}
\vspace{-6mm}
\end{figure*}

\textbf{NBV} has a long history in active perception, and many classical formulations aim to improve global scene representations through multi-view acquisition~\cite{scott2003view}. A common approach uses probabilistic volumetric maps and expected information gain to select the next pose that reveals unseen surfaces and reduces map uncertainty~\cite{aleotti2014global}. Related exploration-oriented planners use receding-horizon view selection to expand known free space and increase coverage~\cite{bircher2016receding, lodel2022look}. These objectives have been demonstrated on real robots for mapping and reconstruction~\cite{gehring2021anymal}.
More recently, NBV has been revisited for learning-based active reconstruction~\cite{wang2024rl}. 
Methods such as GenNBV~\cite{chen2024gennbv} and Auto3R~\cite{shen2025auto3r} learn policies that plan sequences of views to optimize final reconstruction quality over many steps. 
While effective, these systems primarily target long-horizon reconstruction metrics. 
We instead select the next view to obtain a single informative observation of an occluded person under motion constraints.

\textbf{Human-centered active perception} studies how camera motion and viewpoint selection can improve the reliability of estimating people under viewpoint ambiguity and occlusion. A representative direction treats viewpoint as a controllable variable for reducing uncertainty in human pose and motion capture~\cite{gartner2020deep,ci2023proactive}. ActiveMoCap optimizes viewpoints to improve pose estimation quality with a moving camera~\cite{kiciroglu2020activemocap}, and related work learns view-selection policies for informative pose recovery~\cite{gartner2020deep, pirinen2019domes}. Complementarily, ViewActive predicts viewpoint quality from a single image to recommend better viewpoints for perception~\cite{wu2025viewactive}.
Despite these advances, most human-centered viewpoint methods optimize pose-estimation objectives and do not directly target the broader observational usability needed by multiple downstream modules in clutter. In contrast, we study human-centered NBV under severe clutter, where view selection targets an informative next observation while guaranteeing executability on a mobile robot.

\textbf{Occlusion-aware active perception} has long been recognized as a central factor in view utility, especially in cluttered environments where small viewpoint changes can turn a target from visible to largely blocked~\cite{zeng2020view}. Prior NBV work has incorporated explicit visibility reasoning to expose hidden regions during surface acquisition and reconstruction~\cite{dhami2023pred}. More recent learning-based methods model visibility or uncertainty in neural implicit scene representations to guide uncertainty-driven active mapping and view selection~\cite{pan2022activenerf, lee2022uncertainty}. Occlusion-aware view planning also appears in task-driven settings, where the objective is to acquire task-relevant evidence under clutter~\cite{burusa2024semantics,song2025gs}. 
However, many occlusion-aware planners still measure success through coverage, uncertainty, or reconstruction completeness. We instead study single-step target-centric NBV for an occluded person, selecting the next reachable view to immediately improve observation quality for high-reliability tasks such as assessment and triage. \vspace{-4pt}
\section{Methods}\label{sec:Method}

In this section, we present \gls{our_pipeline}, an occlusion-aware next-best-view framework for human-centered active perception on a mobile robot (Fig.~\ref{fig:avo_overview}). It has two stages: (1) \textbf{3D Information Extraction} reconstructs an initial SMPL mesh~\cite{loper2023smpl}, segments the human with SAM~2~\cite{ravi2024sam}, lifts the mask to 3D to extract human points, and aligns mesh and point cloud in the camera frame; and (2) \textbf{Occlusion-aware NBV Generation} samples feasible viewpoints and selects the next view using a occlusion-aware evaluator.

\subsection{3D Information Extraction}

\begin{algorithm}[t]
\caption{Target 3D Information Extraction}
\label{alg:target3d}
\begin{algorithmic}[1]
\State \textbf{Inputs:} RGB image $I$; point cloud $\mathbf{P}=\{p_i\}$; camera intrinsics $\mathbf{K}$
\State \textbf{Outputs:} aligned mesh $M_{\text{aligned}}$; target points $\mathbf{P}_{\text{tgt}}$; background points $\mathbf{P}_{\text{bg}}$; visible part labels $\Pi^\star$
\State \textbf{Initialization:} $\mathbf{P}_{\text{tgt}}\leftarrow\emptyset$, $\mathbf{P}_{\text{bg}}\leftarrow\emptyset$, $\Pi^\star\leftarrow\emptyset$

\State $(b^\star,m^\star,\Pi^\star)\leftarrow \textsc{SATHMRInfer}(I)$
\State $\text{mask}_{2D}\leftarrow \textsc{SAM2Segment}(I,\; b^\star)$

\ForAll{$p\in\mathbf{P}$}
    \State $(u,v)\leftarrow \textsc{ProjectToImage}(p,\mathbf{K})$
    \If{$(u,v)$ is inside image \textbf{and} $\text{mask}_{2D}(u,v)=1$}
        \State $\mathbf{P}_{\text{tgt}}\leftarrow \mathbf{P}_{\text{tgt}}\cup\{p\}$
    \Else
        \State $\mathbf{P}_{\text{bg}}\leftarrow \mathbf{P}_{\text{bg}}\cup\{p\}$
    \EndIf
\EndFor

\If{$\Pi^\star \neq \emptyset$}
    \State $m_{\text{part}}\leftarrow \textsc{ExtractPartSubmesh}(m^\star,\Pi^\star)$
    \State $T_{\text{icp}}\leftarrow \textsc{PointToPlaneICP}(m_{\text{part}},\mathbf{P}_{\text{tgt}})$
\Else
    \State $T_{\text{icp}}\leftarrow \textsc{PointToPlaneICP}(m^\star,\mathbf{P}_{\text{tgt}})$
\EndIf

\State $M_{\text{aligned}}\leftarrow \textsc{ApplyTransform}(m^\star,T_{\text{icp}})$
\State \Return $M_{\text{aligned}},\mathbf{P}_{\text{tgt}},\mathbf{P}_{\text{bg}},\Pi^\star$
\end{algorithmic}
\end{algorithm}

\begin{figure}[t]
  \centering
  \includegraphics[width=1.0\linewidth]{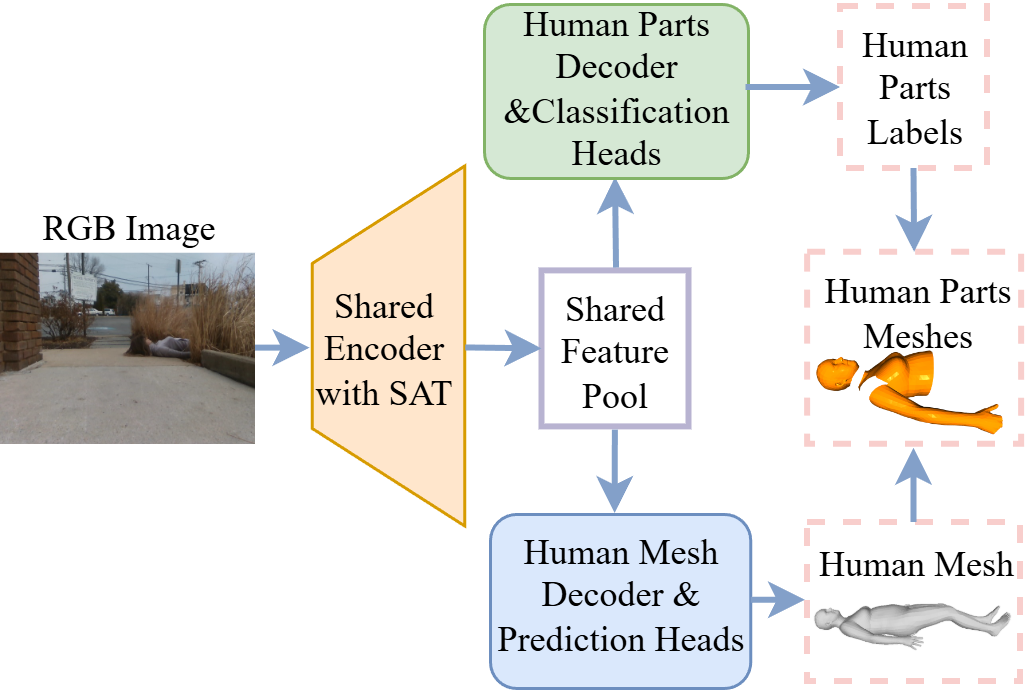}
  \caption{  
    An illustration of our modified SAT-HMR architecture with an additional human-part classification. The input RGB image is encoded by a shared scale-adaptive tokens (SAT) encoder. Two parallel decoders operate on the shared features: a human-mesh decoder that predicts the SMPL mesh and bounding box, and a human-parts decoder with classification heads that assign part labels to human regions. The predicted part cues are used to construct part-specific meshes and enable part-aware mesh-to-point-cloud alignment in Algorithm 1.}
  \label{fig:mesh_model}
\end{figure}

To acquire target 3D information for downstream viewpoint evaluation, \gls{our_pipeline} combines mesh reconstruction, 2D segmentation, and 3D alignment, as summarized in Fig.~\ref{fig:mesh_model} and Algorithm~\ref{alg:target3d}.

For mesh reconstruction, we use SAT-HMR~\cite{su2025sat} for its balance of speed and mesh quality compared to FAST-HMR~\cite{mehraban2026fasthmr} and HSMR~\cite{xia2025reconstructing}. As shown in Fig.~\ref{fig:mesh_model}, we create a SAT-HMR~\cite{su2025sat} variant with a parallel human-parts decoder. Both decoders share SAT encoder features: the mesh branch predicts the target bounding box $b^\star$ and whole SMPL mesh $m^\star$ while the parts branch predicts visible part labels $\Pi^\star$ used to form part-specific mesh subsets. We freeze the SAT-HMR backbone and mesh heads, and train only the part classifier on the CIHP dataset~\cite{gong2018instance}, reaching $71\%$ precision in our setup. 
Given a fixed SMPL vertex-to-part mapping $\phi: V \rightarrow \mathcal{L}$, which maps each mesh vertex $V$ to a body-part label in label list $\mathcal{L}$, we extract the corresponding part submesh
$m_{\text{part}}=\{v\in m^\star \mid \phi(v)\in \Pi^\star\}$ for part-aware alignment.

Because SAT-HMR is appearance-based, its mesh can be poorly localized in the camera frame. We use SAM~2~\cite{ravi2024sam}, prompted by the predicted  bounding box $b^\star$, to obtain a foreground mask, project $\mathbf{P}$ with $\mathbf{K}$, and split points into $\mathbf{P}_{\text{tgt}}$ and $\mathbf{P}_{\text{bg}}$ (Algorithm~\ref{alg:target3d}). We then align the mesh to $\mathbf{P}_{\text{tgt}}$ using point-to-plane ICP on $m_{\text{part}}$ if $\Pi^\star \neq \emptyset$, or on $m^\star$ otherwise. The aligned mesh $M_{\text{aligned}}$ and $\mathbf{P}_{\text{tgt}}$ form the target 3D state used for \gls{nbv} generation.

\subsection{Reachability- and Occlusion-aware NBV Generation} \label{Sec:nbv_pose_generation}

\noindent\textbf{Elevation-Map-Based Viewpoint Generation:} 
While the spherical-shell~\cite{aleotti2014global} and concentric circles generation~\cite{dhami2023pred} perform well in simulation, we observed three critical failure modes when deploying them in the real world with a quadruped robot:  (i) target centroid estimation sometimes places shell centers at unreachable heights, rendering some infeasible viewpoint candidates; (ii) the kinematic coupling between camera height and pitch angle of the robot is ignored; (iii) the depth camera's narrow field of view leaves large regions unobserved, causing target pose candidates to be placed inside obstacles.

To address these limitations, we propose an elevation-map-based viewpoint generation that samples candidates directly on traversable terrain and explicitly models the base-camera kinematic coupling via the robot's rigid-body transformation chain. We leverage the real-time elevation map from Unitree, which has a \SI{60}{\mm} resolution and $128\times128$ grid centered at the robot's base frame.

We first filter out all invalid cells and those exceeding a traversability threshold $h_{step}=0.15\,m$. Then, $\mathbf{T}_i^{\text{base}}$ are uniformly and randomly sampled in the map. Each position is elevated by $h=0.3\,m$ above the terrain surface according to the robot's standing height. Then the robot is oriented towards the target, and camera poses can be derived with a fixed-rigid-body transform $\mathbf{T}_{\text{cam}}^{\text{base}}\in SE(3)$ and a local y-axis rotation $\mathbf{T}_y(\alpha)\in SE(3), \alpha\in[-0.75,\,0.75]\,rad$ to cover robot's pitch working range:
\vspace{-1pt}
\begin{equation}
\mathbf{T}_i^{\text{cam}}(\alpha) = \mathbf{T}_{\text{base}}^{\text{cam}}\cdot\mathbf{T}_y(\alpha)\cdot\mathbf{T}_i^{\text{base}},
\end{equation}\vspace{-1pt}where $i = 1,2,3,...,M$. We select $M=100$, which is sufficient to cover the traversable cells in the forward-facing half of the map, as occluded and out-of-view cells are excluded from sampling. Additionally, for each pose $\mathbf{T}_i^{\text{base}}$, we uniformly sample 10 different angles within the range defined by $\alpha$.

\newlength{\LeftColW}
\setlength{\LeftColW}{8mm}

\newlength{\RowGap}
\setlength{\RowGap}{3.4pt}

\newlength{\RowGapLast}
\setlength{\RowGapLast}{4.5pt}

\newlength{\WhiteExtraLeft}
\setlength{\WhiteExtraLeft}{7mm}

\newlength{\LabelStripH}
\setlength{\LabelStripH}{18pt}

\newlength{\LabelStripHLast}
\setlength{\LabelStripHLast}{5pt}

\newlength{\HeadPadX}
\setlength{\HeadPadX}{18.25mm}

\newlength{\HeadH}
\setlength{\HeadH}{7mm}

\newlength{\HeadMinW}
\setlength{\HeadMinW}{7mm}

\newcommand{\HeadBox}[2][headergray]{%
  \colorbox{#1}{%
    \parbox[c][\HeadH][c]{\dimexpr 2\HeadPadX + \HeadMinW \relax}{%
      \centering #2%
    }%
  }%
}

\newcommand{\BottomWhiteStrip}{%
  \rlap{%
    \hspace*{-\WhiteExtraLeft}%
    \raisebox{-\LabelStripH}[0pt][0pt]{%
      \color{white}%
      \rule{\dimexpr\LeftColW+\WhiteExtraLeft\relax}{\LabelStripH}%
    }%
  }%
}

\newcommand{\BottomWhiteStripLast}{%
  \rlap{%
    \hspace*{-\WhiteExtraLeft}%
    \raisebox{-\LabelStripHLast}[0pt][0pt]{%
      \color{white}%
      \rule{\dimexpr\LeftColW+\WhiteExtraLeft\relax}{\LabelStripHLast}%
    }%
  }%
}

\newcommand{\LeftLabelCell}[2]{
  \cellcolor{#1}%
  \smash{%
    \raisebox{15mm}{%
      \rotatebox[origin=c]{90}{\fontsize{9pt}{10pt}\selectfont #2}%
    }%
  }%
}

\begin{figure*}[!t]
  \centering
  \setlength{\tabcolsep}{0pt}
  \renewcommand{\arraystretch}{1}
  \setlength{\fboxsep}{0pt}

  \resizebox{\textwidth}{!}{%
    \begin{tabular}{@{}>{\centering\arraybackslash}m{\LeftColW}@{\hspace{0.3mm}}cccc@{}}

      & \HeadBox[headergray]{Initial Overall View}
      & \HeadBox[initial_camera_view]{Initial Camera View}
      & \HeadBox[headergray]{Best Overall View}
      & \HeadBox[best_camera_view]{Best Camera View} \\[4pt]

      \LeftLabelCell{pastelblue}{Indoor Simulation}\BottomWhiteStrip &
      \cellbox{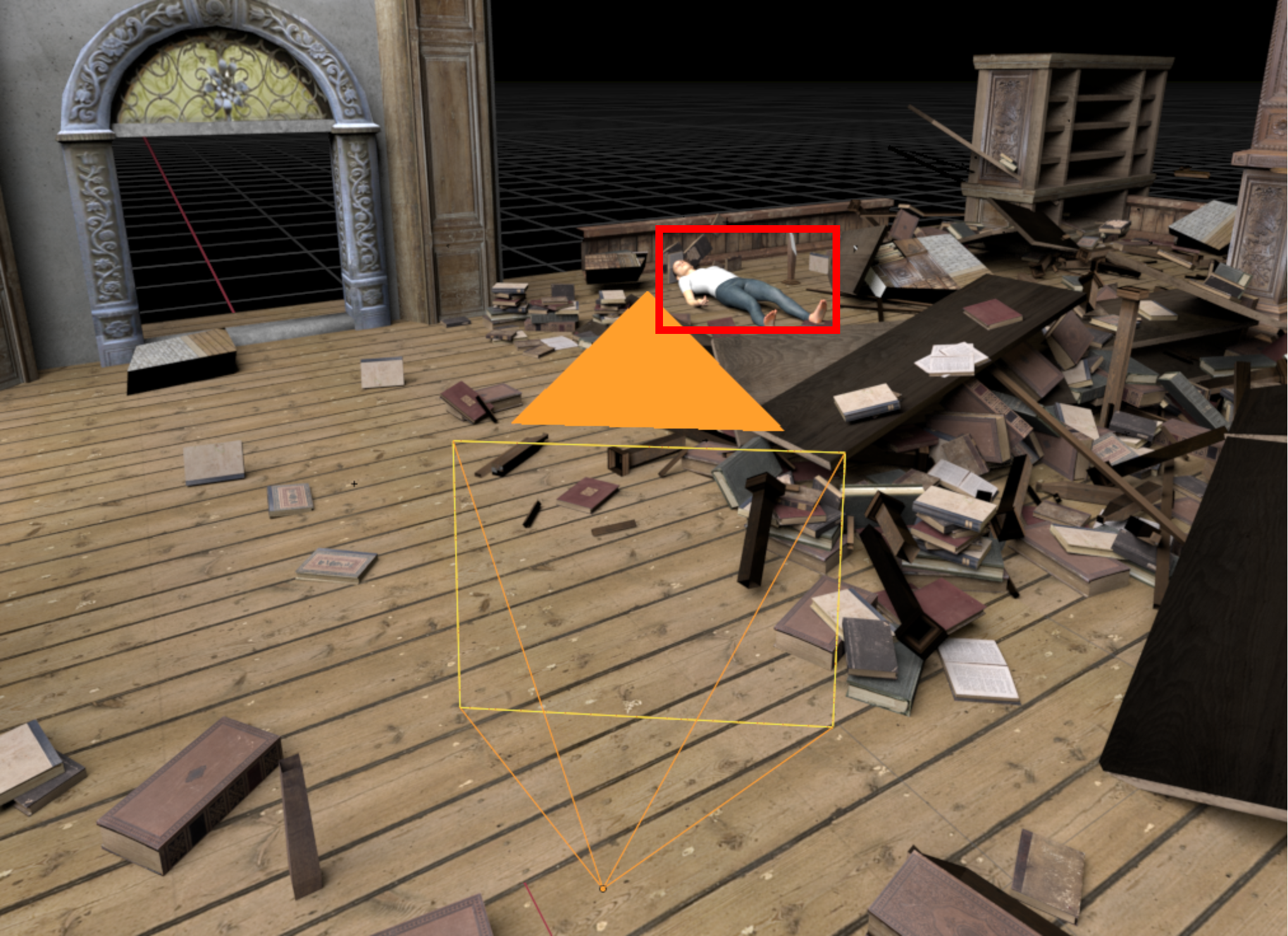} &
      \cellbox{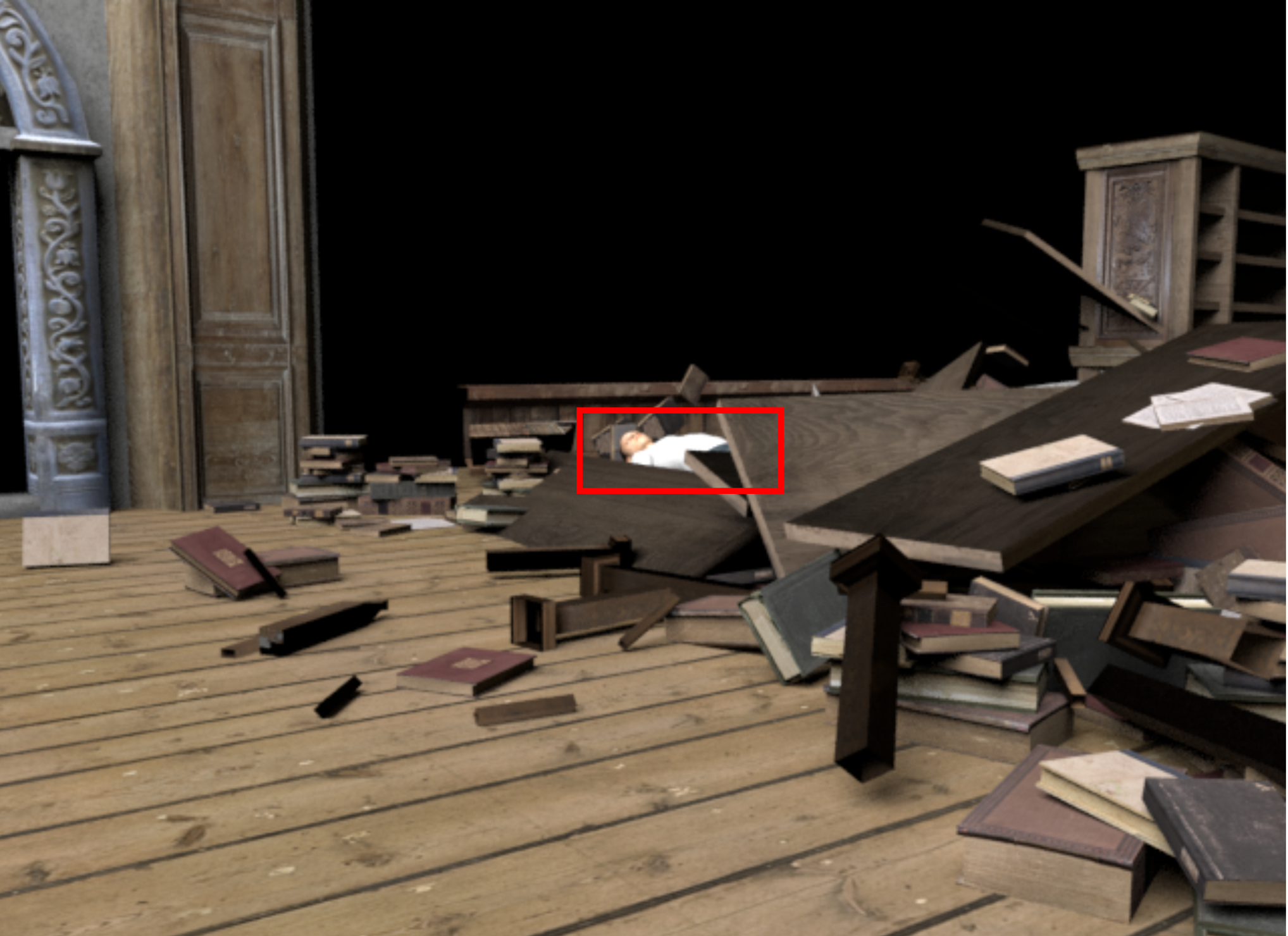} &
      \cellbox{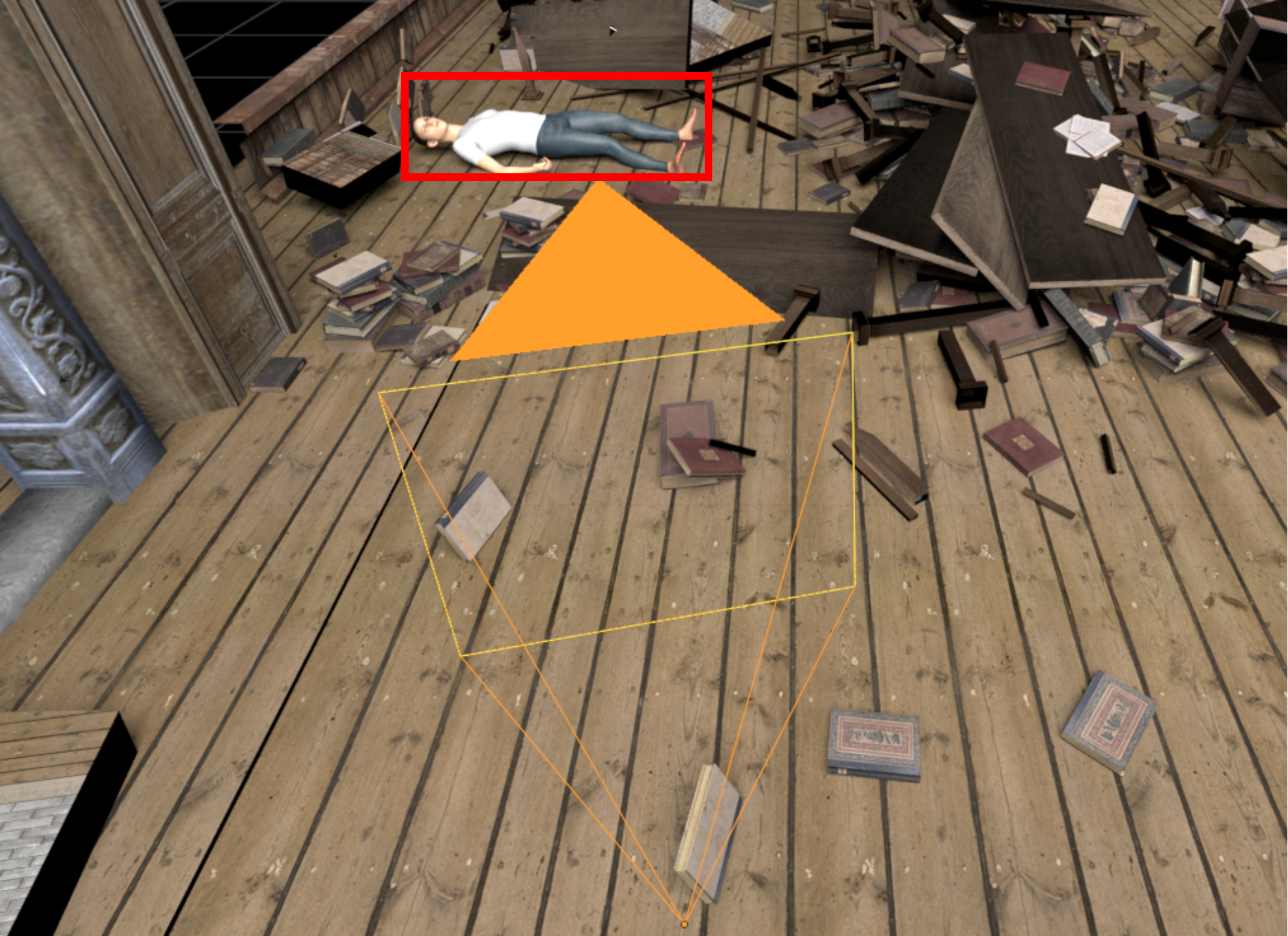} &
      \cellbox{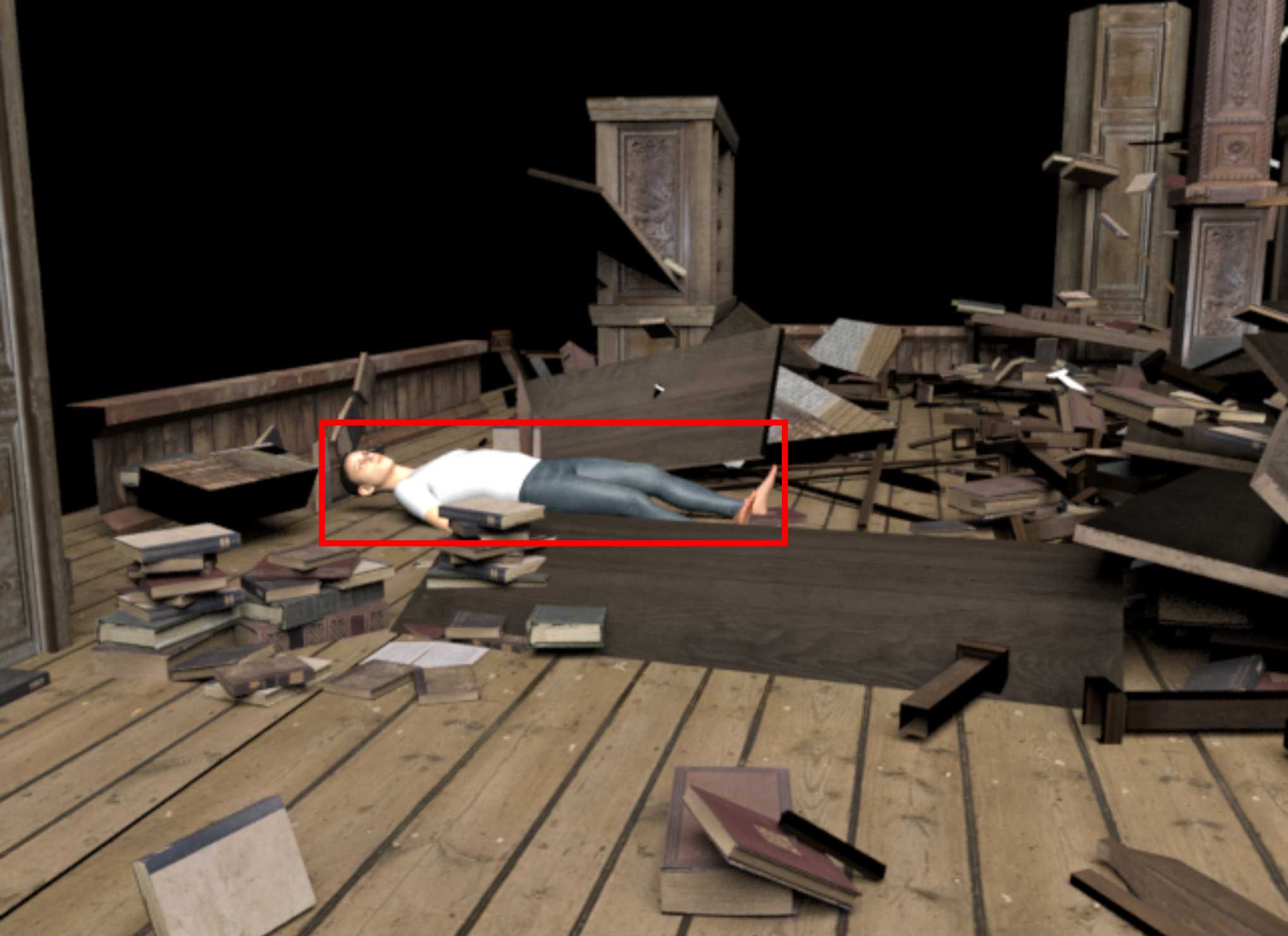} \\[-\RowGap]

      \LeftLabelCell{pastelpink}{Outdoor Simulation}\BottomWhiteStrip &
      \cellbox{images/Experiment/experiment_2_1} &
      \cellbox{images/Experiment/experiment_2_2} &
      \cellbox{images/Experiment/experiment_2_3} &
      \cellbox{images/Experiment/experiment_2_4} \\[-\RowGap]

      \LeftLabelCell{pastelpurple}{Indoor Real-world}\BottomWhiteStrip &
      \cellbox{images/Experiment/experiment_3_1} &
      \cellbox{images/Experiment/experiment_3_2} &
      \cellbox{images/Experiment/experiment_3_3} &
      \cellbox{images/Experiment/experiment_3_4} \\[-\RowGap]

      \LeftLabelCell{pastelyellow}{Outdoor Real-world}\BottomWhiteStripLast &
      \cellbox{images/Experiment/experiment_4_1} &
      \cellbox{images/Experiment/experiment_4_2} &
      \cellbox{images/Experiment/experiment_4_3} &
      \cellbox{images/Experiment/experiment_4_4}

    \end{tabular}%
  }

  \caption{
    \textbf{Overview of simulation and real-world environments.}
    From top to bottom, the rows correspond to indoor simulation, outdoor simulation, indoor real-world, and outdoor real-world. From left to right, the columns show the initial overall view, the initial camera view, the best overall view, and the best camera view. The red bounding box indicates the human’s location. In simulation environments, the yellow box denotes the camera position.\vspace{-6pt} }
  \label{fig:qual_4x4_wide}
  \vspace{-7mm}
\end{figure*}

\noindent\textbf{Viewpoint evaluator.}
For each candidate viewpoint $i$, we transform the scene points ($\mathbf{P}_{\text{tgt}}$ and $\mathbf{P}_{\text{bg}}$) into the candidate camera frame and project both the aligned mesh vertices and scene points onto the image plane.
The viewpoint is scored by a weighted sum of three image-space terms that capture in-frame completeness, target scale, and occlusion:
\begin{equation}
\label{eq:score_compact}
S_{\text{total}}^{i}= w_v S_v^{i}+ w_a S_a^{i}+ w_o S_o^{i}, \qquad w_v+w_a+w_o=1,
\end{equation}
\begin{equation}
\label{eq:components_compact}
S_v^{i}=\frac{n^{i}_{\text{in}}}{n_m},\qquad
S_a^{i}=\frac{n^{i}_{\text{in}}}{n_I},\qquad
S_o^{i}=1-\frac{n^{i}_{\text{occ}}}{n^{i}_{\text{in}}}.
\end{equation}
Here, $n_m$ is the number of mesh vertices, $n_I$ is the number of image pixels, $n^{i}_{\text{in}}$ counts mesh vertices that project inside the image bounds, and $n^{i}_{\text{occ}}$ counts those vertices that are occluded by the scene geometry using a depth test between the projected mesh and point cloud.

Each term targets a failure mode that commonly degrades reconstruction from a single view.
The visibility term $S_v$ discourages viewpoints where parts of the body fall outside the image boundary, which leads to incomplete observations.
The area term $S_a$ favors viewpoints where the target occupies a larger portion of the image, improving pixel-level detail for keypoints, masks, and mesh regression.
The occlusion term $S_o$ penalizes views where the target is blocked by scene geometry, since occluded observations often cause unstable detections and inaccurate alignment.
We select the next view as $i^\star=\arg\max_i S_{\text{total}}^{i}$.
The weights $(w_v,w_a,w_o)$ are tuned in Sec.~\ref{sec:param_tuning}.
\section{Experiments and Results}

To assess \gls{our_pipeline}, we perform experiments compared with baseline methods on both simulation and real-world environments.
The simulation setup provides precise camera control, ground-truth 3D geometry, and occlusion-free reference views, enabling weight tuning for $(w_v, w_a, w_o)$ and full pipeline testing. In the real world, we deploy \gls{our_pipeline} to a quadruped robot and execute experiments with a mannequin.

\subsection{Experiment Setup}
\vspace{-3pt}
\noindent\textbf{Simulation \& Real World:}
We evaluate \gls{our_pipeline} in simulation environments built using Blender 4.5, and on a Unitree Go2 quadruped. Fig.~\ref{fig:qual_4x4_wide} shows an overview of testing environments.

In simulation, we place SMPL humans in the DISC post-disaster scenes~\cite{jeon2019disc} and custom scenes, randomizing subject pose and the initial camera view to ensure partial body visibility. We render paired RGB and point cloud observations, record camera intrinsics/extrinsics, and export the ground-truth mesh. We include indoor and outdoor settings, each with 15 sub-environments, and sample 2500 occluded initial views. From each start, we run an iterative render--evaluate--move loop: sample candidate viewpoints, score them with our evaluator, and move to the selected next view.

For real-world experiments, a Unitree Go2 Edu carries a Unitree L1 LiDAR and a RealSense D435i RGB-D camera. An NVIDIA Jetson Orin (Ubuntu 20.04, ROS2 Foxy) handles viewpoint generation and navigation, while segmentation and mesh reconstruction/registration run on a laptop (AMD Ryzen 9 7945HX + NVIDIA RTX 4090 Laptop GPU) via 5 GHz Wi-Fi. We evaluate 6 scenarios (3 indoor, 3 outdoor)  with varied obstacles and poses; each run starts from a fixed initial pose (no prior map) and navigates to the selected best viewpoint. We run 10 trials for each method in each scenario, starting from the same initial pose.

\begin{figure}[htbp]
  \centering
  \includegraphics[width=\linewidth]{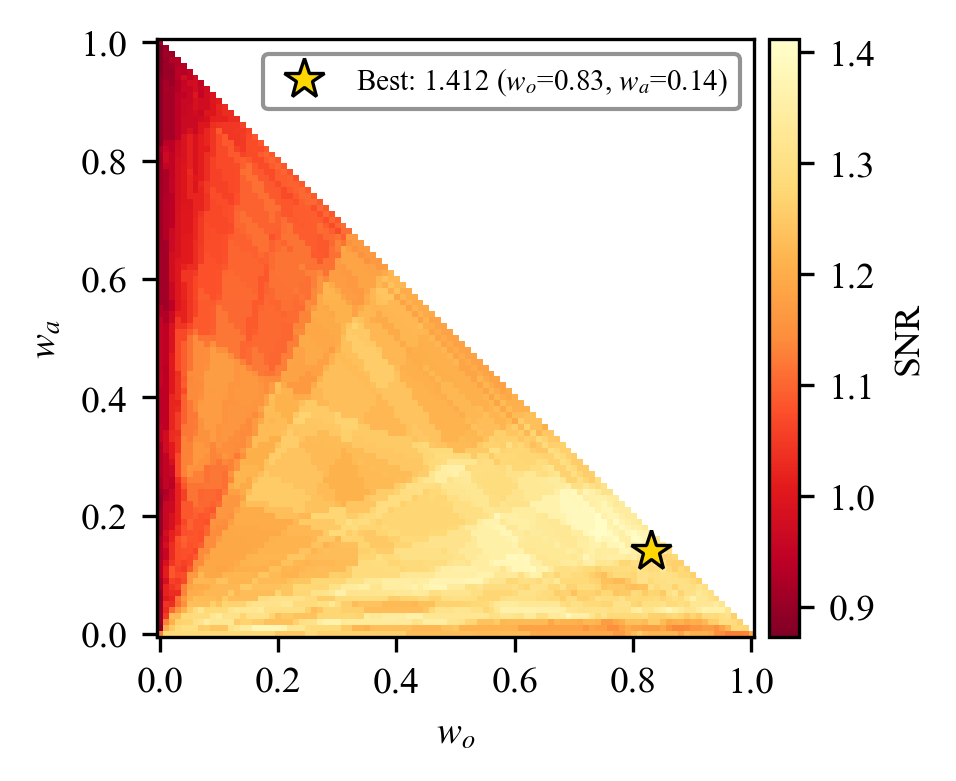}
  \caption{\textbf{SNR heatmap of evaluator weight sweep testing.} The x-axis is the occlusion weight $w_o$ and the y-axis is the target-area weight $w_a$, with the remaining visibility weight $w_v = 1 - w_o - w_a$. The star marks the best-performing weight setting.\vspace{-8pt}}
  \label{fig:weights_heatmap}
\end{figure}

\noindent\textbf{Baselines:} We compare two baselines for 3D target-centric next best view planning of an occluded human. \glsentryshort{volnbv}~\cite{aleotti2014global} selects views by occupancy grid information gain from observed partial geometry. \glsentryshort{prednbv}\cite{dhami2023pred} first completes the target into a predicted point cloud, then selects views by prediction-based gain. In simulation, all methods sample candidates on a spherical shell centered at the target with radii from \SI{2}{m} to \SI{5}{m} (step size \SI{0.5}{m}). On each shell, we uniformly sample 100 candidates pointing toward the target, which is denser than the original Pred-NBV sampling~\cite{dhami2023pred}. In real-world experiments, all methods use our elevation-map-based viewpoint generation (Sec.~\ref{Sec:nbv_pose_generation}) to ensure physical feasibility. Thus, the three methods differ only in their scoring functions.

\noindent\textbf{Metrics:} We report success rate, normalized target area, and keypoint visibility. Success rate is the fraction of trials where the new view yields a valid human detection. Normalized target area is $A = M/(H \times W)$, where $M$ is the target mask pixel count and $H \times W$ is the image resolution, reflecting how large the person appears in the image. Keypoint visibility is $R_{\text{vis}} = n_{\text{vis}}/n_{\text{kp}}$, where $n_{\text{vis}}$ is the number of visible 2D keypoints detected by RTMPose~\cite{jiang2023rtmpose}, reflecting the completeness of the target. If detection fails, we set $A=0$ and $R_{\text{vis}}=0$. For weight tuning, we use \glsentrylong{snr} $\mathrm{SNR}=\mu(R_{\text{vis}})/\sigma(R_{\text{vis}})$ over randomized trials. In ablations, we also report mean per-vertex position error (MPVPE), the mean Euclidean distance between estimated and ground-truth mesh vertices.

\subsection{Weight Tuning}
\vspace{-3pt}\label{sec:param_tuning}
We tune the weighting coefficients in Eq.~\eqref{eq:score_compact} by a grid sweep in simulation.
For each trial, we initialize the camera at a randomly sampled pose, following Sec.~\ref{Sec:nbv_pose_generation}, render an RGB image and point cloud, and run our viewpoint generation to sample candidate viewpoints, excluding candidates that are visually invalid. 
For every candidate, we run RTMPose~\cite{jiang2023rtmpose} on the rendered image to compute the keypoint visibility.
Independently, for each candidate, we compute the evaluator score $S_{\text{total}}$ in Eq.~\eqref{eq:score_compact} using a weight triple $(w_v,w_a,w_o)$ constrained by $w_v+w_a+w_o=1$, and select the candidate that maximizes $S_{\text{total}}$ as our predicted best view. 
We record each weight triplet by the keypoint visibility achieved at the predicted best view and report the SNR over 2500 randomized trials.
Fig.~\ref{fig:weights_heatmap} shows the sweep over $(w_o,w_a)$, with $w_v$ recovered as $w_v = 1 - w_o - w_a$. It demonstrates that the optimum occurs at $(w_v,w_a,w_o)=(0.03,\,0.14,\,0.83)$, indicating that strong occlusion awareness is most important for selecting informative next views in our scenarios, while target-area and in-frame visibility provide complementary but smaller gains.

\subsection{Simulation Results}
\label{sec:simulation_experiments}

\begin{figure}[!h]
  \centering
  \includegraphics[width=0.9\linewidth]{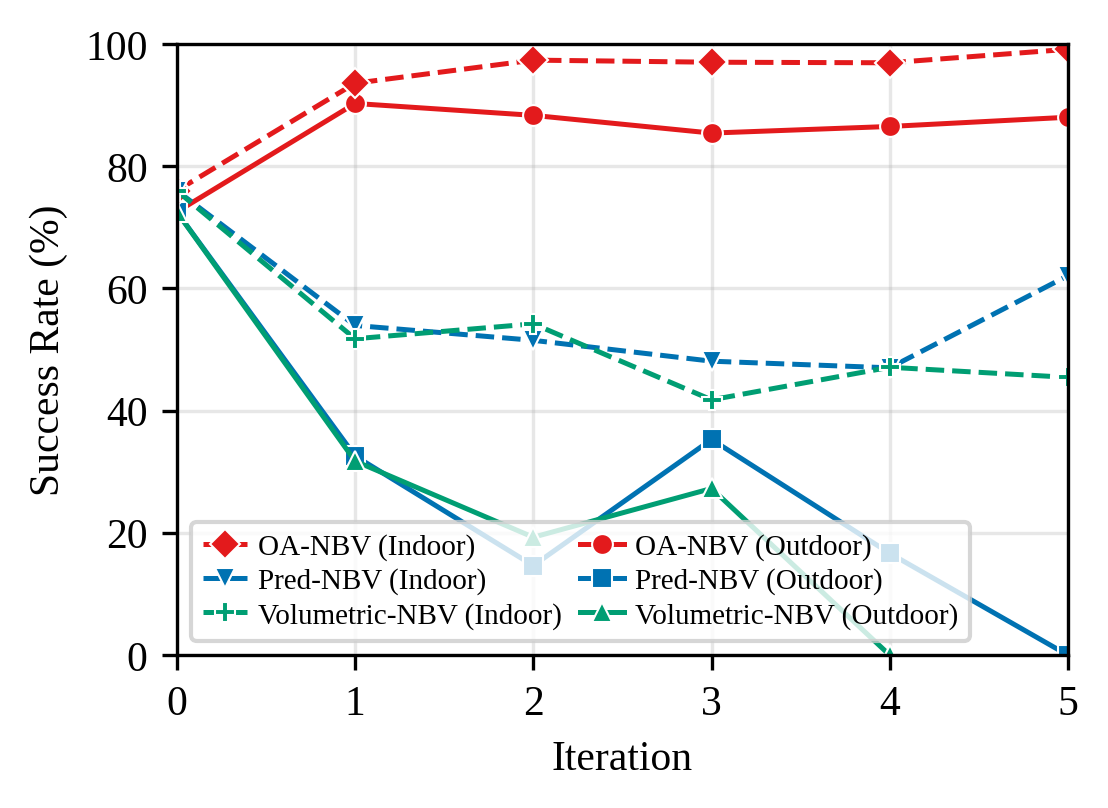}
    \caption{\textbf{Simulation success rate over iterative NBV.} Success is evaluated on the newly acquired view at each step. \gls{our_pipeline} maintains the highest success across iterations in both indoor and outdoor scenes, while Pred-NBV~\cite{dhami2023pred} and Volumetric-NBV~\cite{aleotti2014global} degrade, especially outdoors.}
  \label{fig:iteration_successful_rate}
\end{figure}

\vspace{-3pt}
Using the weights selected in Sec.~\ref{sec:param_tuning}, we first compare \gls{our_pipeline} with \glsentryshort{prednbv}~\cite{dhami2023pred} and \glsentryshort{volnbv}~\cite{aleotti2014global} by reporting success rate over iterative NBV steps, where a trial is marked successful if SAT-HMR\cite{su2025sat} returns a valid bounding box with confidence above $0.3$. Fig.~\ref{fig:iteration_successful_rate} reports the success rate versus iteration.
Across both environments, \gls{our_pipeline} maintains consistently high success, indicating that explicitly accounting for occlusion provides reliable viewpoint selection.
In contrast, \glsentryshort{prednbv} and \glsentryshort{volnbv} degrade markedly over later iterations, especially outdoors: although they may select views that are beneficial under geometric or coverage criteria, the chosen viewpoints are more often blocked by surrounding structures when occlusion is not modeled, leading to missed detections.
Finally, indoor success rates are generally higher than outdoor success rates in our setup, since the outdoor scenes contain more clutter and large occluders overall, making robust human detection more difficult.

\begin{figure}[!h]
  \centering
  \includegraphics[width=0.9\linewidth]{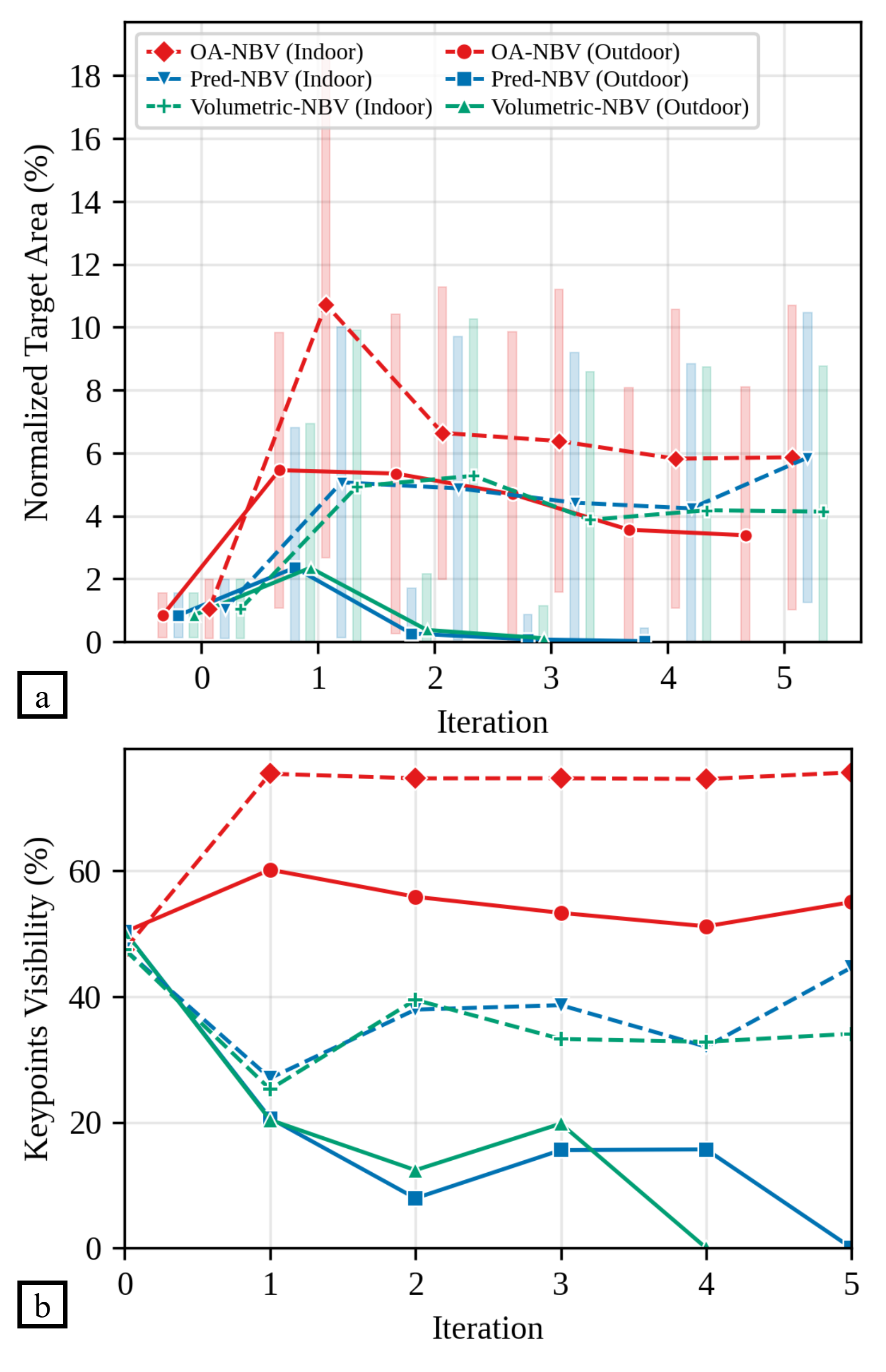}
  \caption{\textbf{Simulation view-quality over iterative NBV.} (a) Normalized target area $A$. (b) Keypoint visibility $R_{\text{vis}}$. Curves show mean over trials. \gls{our_pipeline} maintains higher $A$ and $R_{\text{vis}}$ across iterations, while \glsentryshort{prednbv}~\cite{dhami2023pred} and \glsentryshort{volnbv}~\cite{aleotti2014global} degrade. \vspace{-8pt}}
  \label{fig:simulation_iterations}
\end{figure}

We also compare how view quality changes over iterative \gls{nbv} steps in simulation.
To characterize view quality, we report normalized target area $A$ and keypoint visibility $R_{\text{vis}}$ in Fig.~\ref{fig:simulation_iterations}. Higher values in both metrics indicate more informative viewpoints for downstream reconstruction.
As shown in Fig.~\ref{fig:simulation_iterations}, \gls{our_pipeline} consistently achieves a larger target area and higher keypoint visibility across iterations in both indoor and outdoor scenes.
In particular, the sharp improvement from iteration~0 to~1 indicates that a single \gls{nbv} update already moves the camera toward a more informative view.
In the indoor setting, $A$ drops after iteration~1 because the first move often reaches a near-optimal view. Later steps explore alternative reachable viewpoints, which remain useful but may produce a smaller projected target area. This reflects diminishing returns as nearby candidates become similar.
In contrast, \glsentryshort{prednbv} and \glsentryshort{volnbv} often drift toward more occluded viewpoints over later iterations. This reduces keypoint visibility and target area, especially outdoors, where clutter and large occluders are common. \vspace{-5pt}

\begin{table}[b]
\centering
\caption{\textbf{Best simulation performance across NBV iterations in indoor and outdoor environments} ($N=2500$, units: \%).
We report the peak value over iterations~1--5 for each metric.
\gls{our_pipeline} achieves the highest peaks across metrics in both settings.}
\label{tab:main_results_simulation}
\small
\begin{tabular}{ll|ccc}
\toprule
& Method & \makecell{Keypoints\\visibility} & \makecell{Normalized\\target area} & \makecell{Success\\rate} \\
\midrule
\multirow{3}{*}{\rotatebox{90}{Indoor}}
& Volumetric-NBV & 39.5 & 5.3  & 54.2 \\
& Pred-NBV       & 44.6 & 5.9  & 62.1 \\
& \textbf{\gls{our_pipeline}} & \textbf{75.7} & \textbf{10.7} & \textbf{99.1} \\
\midrule
\multirow{3}{*}{\rotatebox{90}{Outdoor}}
& Volumetric-NBV & 20.4 & 2.4 & 31.8 \\
& Pred-NBV       & 20.6 & 2.4 & 32.6 \\
& \textbf{\gls{our_pipeline}} & \textbf{60.2} & \textbf{5.5} & \textbf{90.3} \\
\bottomrule
\end{tabular}
\end{table}

Table~\ref{tab:main_results_simulation} summarizes the best achievable performance across all iterations for each method.
\gls{our_pipeline} attains the strongest peaks in both indoor and outdoor scenes across all metrics, with a larger margin outdoors under heavier occlusion.
Moreover, the highest view-quality metrics are typically achieved after the first \gls{nbv} update, suggesting diminishing returns from additional iterations in our setting.
Based on this observation, we use a single \gls{nbv} step in real-world experiments.
This choice also supports a fair comparison, because multi-step rollouts amplify differences in detection success under occlusion and can disproportionately reduce the reported performance of \glsentryshort{prednbv} and \glsentryshort{volnbv}.
\vspace{-8pt}
\subsection{Real World Results}
We evaluate \gls{our_pipeline} on a Unitree Go2 in indoor and outdoor scenes using the same baselines and metrics as in simulation. Based on the diminishing returns observed in Sec.~\ref{sec:simulation_experiments}, we apply a single \gls{nbv} step in all real-world experiments. Table~II reports the quantitative results over $N=30$ trials per setting. \gls{our_pipeline} achieves the highest performance across all three metrics in both environments. In indoor experiments, it attains \SI{68.9}{\percent} keypoint visibility, \SI{7.0}{\percent} normalized target area, and an over \SI{90}{\percent} success rate. In outdoor experiments, \gls{our_pipeline} maintains an over \SI{90}{\percent} success rate with \SI{71.2}{\percent} keypoint visibility and \SI{6.7}{\percent} target area, while the baseline methods exhibit substantially lower success and visibility. These results are consistent with the findings observed in simulation and confirm the effectiveness of \gls{our_pipeline} in real-world cluttered environments.

\begin{table}[t]
\centering
\caption{\textbf{Real-world performance after a single NBV update in indoor and outdoor environments} ($N=30$, units: \%). \gls{our_pipeline} still shows the best performance, consistent with simulation.}
\label{tab:main_results_realworld}
\small
\begin{tabular}{ll|ccc}
\toprule
& Method & \makecell{Keypoints\\visibility} & \makecell{Normalized\\target area} & \makecell{Success\\rate} \\
\midrule
\multirow{3}{*}{\rotatebox{90}{Indoor}}
& Volumetric-NBV & 10.9 & 1.5  & 10.0 \\
& Pred-NBV       & 29.3 & 1.9  & 43.3 \\
& \textbf{\gls{our_pipeline}} & \textbf{68.9} & \textbf{7.0} & \textbf{90.0} \\
\midrule
\multirow{3}{*}{\rotatebox{90}{Outdoor}}
& Volumetric-NBV & 12.2 & 2.1 & 13.3 \\
& Pred-NBV       & 45.1 & 3.7 & 16.7 \\
& \textbf{\gls{our_pipeline}} & \textbf{71.2} & \textbf{6.7} & \textbf{96.7} \\
\bottomrule
\end{tabular}
\end{table}
\section{Discussion}
\subsection{Ablation study}
\begin{figure}[!h]
  \centering
  \includegraphics[width=0.8\linewidth]{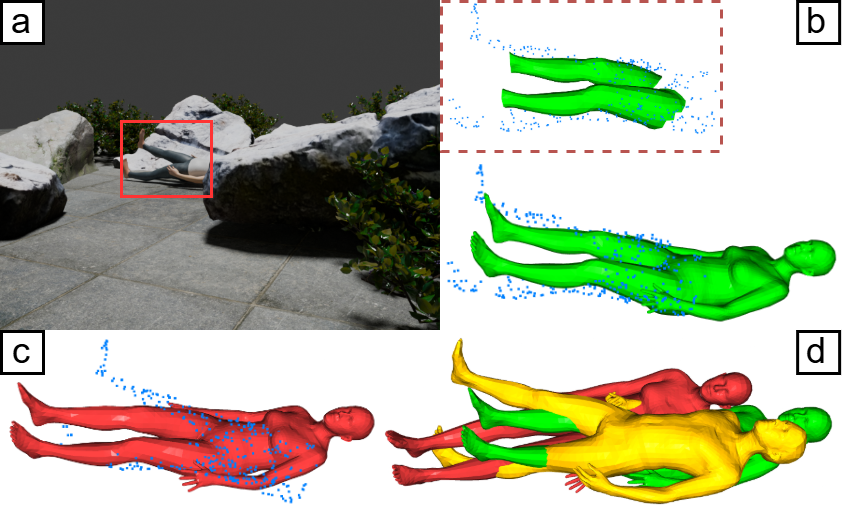}
  \caption{\textbf{Mesh-alignment ablation comparing part-mesh and full-mesh registration under partial observations.} (a) Outdoor scene with a heavily occluded target. (b) Part-mesh registration aligns the whole mesh (green) to visible target points (blue). (c) Full-mesh registration converges to an incorrect pose (red). (d) Overlay comparison: part-mesh vs. full-mesh vs. ground truth (gold).}
  \label{fig:ablation_mesh}
\end{figure}

\vspace{-3pt}
\noindent\textbf{Part-mesh alignment vs. full-mesh alignment:}
Our alignment module registers a human mesh to the observed target point cloud to estimate the target state for downstream planning. When the observation is heavily occluded, the point cloud is incomplete, and full-body registration becomes ambiguous. Similar local geometry across different body parts can create spurious correspondences, pulling the optimizer toward the wrong region and corrupting the pose estimate, even leading to failure in viewpoint prediction. To reduce this ambiguity, we instead register a part mesh that corresponds to the visible region. To quantify the effect, we compare part-mesh registration against full-mesh registration in the outdoor simulation setting, where occlusion and clutter are most challenging. We run $500$ randomized trials and evaluate alignment accuracy using MPVPE. As shown in Fig.~\ref{fig:ablation_mesh}, part-mesh alignment consistently improves registration quality, changing the average MPVPE from \SI{386.36}{\mm} to \SI{233.79}{\mm}, reducing by \SI{39.3}{\percent} on average compared to full-mesh alignment. This result supports our design choice to use part-specific geometry for robust alignment under partial observations.

\begin{figure}[htbp]
  \centering
  \includegraphics[width=1\linewidth]{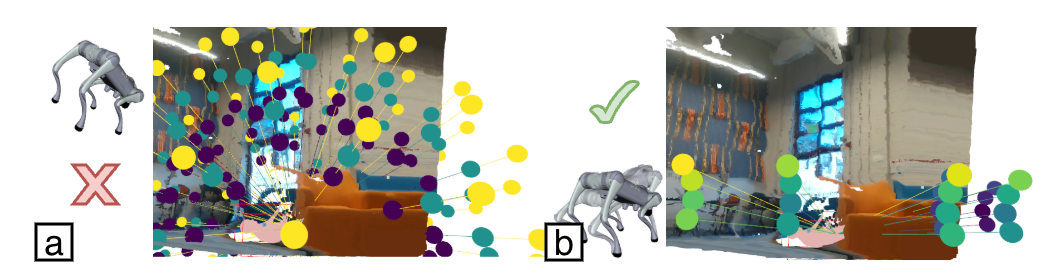}
\caption{Comparison of viewpoint generation strategies in a real-world trial. (a) The spherical-shell method~\cite{aleotti2014global} places candidates at unreachable heights or inside obstacles. (b) Our elevation-map-based method samples candidates directly on traversable terrain, ensuring kinematically feasible and collision-free viewpoints. We only show a local region here for clarity.}
  \label{fig:compare_pose_generator}
\end{figure}

\noindent\textbf{Elevation-map-based viewpoint generation vs. Spherical Shell viewpoint generation:} As described in~\ref{Sec:nbv_pose_generation} and Fig.~\ref{fig:compare_pose_generator}, the spherical shell viewpoint generation faces several practical challenges when deployed on a legged robot. We conducted an ablation study comparing these two approaches across 10 trials spanning open scenarios and cluttered environments with identical pipeline components. In these experiments, our proposed generation succeeded all 10 times, while the spherical shell generation succeeded only 3 times. In 5 failures, candidate viewpoints fell inside obstacles and were physically unreachable. In the remaining two, viewpoints were situated above low-lying obstacles. After projecting the coordinates on the ground, the robot ended up behind the obstacle without sight of the target. 

In contrast, by generating all candidate viewpoints based on the elevation map, our generation ensures physical feasibility and correct camera position in the world frame for the viewpoint evaluator, avoiding both failure modes. Notably, the spherical shell generation occasionally succeeds in open scenarios despite inaccurate camera position estimates, suggesting that the absence of obstacles partially compensates for its geometric limitations. However, this robustness breaks down in cluttered environments, where precise viewpoint placement becomes critical for reliable target observation.

\vspace{-8pt}
\subsection{Limitations and future work.}
\vspace{-3pt}
Our real-world performance is sensitive to outdoor illumination. Even with the same hardware and configuration, darker lighting conditions occasionally fail the person detector, preventing the pipeline from running end-to-end.
Second, OA-NBV targets single-step viewpoint selection for immediate observation quality, rather than multi-view reconstruction. Tasks requiring complete surface coverage may need viewpoints that reveal different unseen regions, which our occlusion-aware scoring does not prioritize.
In future work, we plan to (i) improve robustness for challenging sensing conditions such as low light, smoke, and dust, (ii) use single-step viewpoint selection as a first stage, followed by optional multi-view refinement for downstream tasks requiring higher accuracy, such as injury assessment and manipulation planning, and (iii) extend OA-NBV to multi-person scenarios with target prioritization and identity tracking.
\vspace{-3pt} 
\section{Conclusion}

We presented OA-NBV, an occlusion-aware next-best-view framework that enables a mobile robot to select a traversable viewpoint that is immediately usable for human-centered perception in clutter. OA-NBV closes the loop between perception and motion by estimating a target 3D hypothesis from partial observations and scoring feasible candidate views with a target-centric model that accounts for visibility, scale, and completeness.

Across extensive experiments in simulation and on a Unitree Go2 Edu, OA-NBV improves both robustness and observation quality compared with representative volumetric and prediction-guided baselines. In both simulation and real-world testing, OA-NBV achieves over \SI{90}{\percent} success across settings. It also improves the normalized target area by at least \SI{81}{\percent} and keypoint visibility by at least \SI{58}{\percent} compared to the strongest baseline. These results indicate that OA-NBV is a practical autonomous view-selection module for field robots supporting human-centered tasks such as search, triage, and disaster response, especially when medical personnel and robot operators are limited.

\bibliography{bibliography}
\bibliographystyle{ieeetr}
\end{document}